\documentclass[letterpaper, 10 pt, conference]{ieeeconf}  

\IEEEoverridecommandlockouts                              

\usepackage{latexstyle}

\usepackage[noadjust]{cite}
\usepackage{color}

\newif\ifshowchanges
\showchangestrue
\showchangesfalse

\newenvironment{newremoval}{%
   \color{red}
}{}
\newenvironment{newaddition}{%
   \color{Paired6qual4}
}{}
\newcommand{\revisionchange}[2]{%
    \ifshowchanges%
        \ifx&#1&%
            \begin{newaddition}%
                #2%
            \end{newaddition}%
        \else%
            \begin{newremoval}%
                #1%
            \end{newremoval}%
            \begin{newaddition}%
                #2%
            \end{newaddition}%
        \fi%
    \else%
        #2%
    \fi%
}%

\title{\LARGE \bf COP: Control \& Observability-aware Planning}%

\author{Christoph Böhm$^{1}$, Pascal Brault$^{2}$, Quentin Delamare$^{2}$, Paolo Robuffo Giordano$^{3}$, and Stephan Weiss$^{1}$
\thanks{$^{1}$C. Böhm and S. Weiss are with the Control of Networked Systems Group at the University of Klagenfurt, Austria. {\tt\scriptsize email: \{firstname.lastname\}@ieee.org}}%
\thanks{$^{2}$P. Brault and Q. Delamare are with ENS, Univ Rennes, Inria, IRISA, Campus de Beaulieu,
35042 Rennes Cedex, France. {\tt\scriptsize email: pascal.brault@irisa.fr, quentin.delamare@irisa.fr}}%
\thanks{$^{3}$P. Robuffo Giordano is with CNRS, Univ Rennes, Inria, IRISA, Campus de Beaulieu,
35042 Rennes Cedex, France. {\tt\scriptsize email: prg@irisa.fr}}%
\thanks{This work was partially supported by the project  ANR-20-CE33-0003 ``CAMP'' and by the Austrian Ministry for Transport, Innovation and Technology (BMVIT)
under the grant agreement 878661 (SCAMPI)}
\thanks{{\textbf{Accepted January/2022 for ICRA 2022, DOI follows ASAP~\copyright IEEE.}}}
}%

\begin{document}

\maketitle
\global\csname @topnum\endcsname 0
\global\csname @botnum\endcsname 0
\thispagestyle{empty}
\pagestyle{empty}


\begin{abstract}
    In this research, we aim to answer the question: \emph{How to combine Closed-Loop State and Input Sensitivity-based with Observability-aware trajectory planning?} These \emph{possibly opposite} optimization objectives can be used to improve trajectory control tracking and, at the same time, estimation performance.
    
    Our proposed novel \revisionchange{Control and Observability-aware Planning (COP)}{\gls{cop}} framework is the first that uses these \emph{possibly opposing} objectives in a Single-Objective Optimization Problem (SOOP) based on the Augmented Weighted Tchebycheff method to perform the balancing of them and generation of Bézier curve-based trajectories. Statistically relevant simulations for a 3D quadrotor \gls{uav} case study produce results that support our claims and show the negative correlation between both objectives. We were able to reduce the positional mean integral error norm as well as the estimation uncertainty with the same trajectory to comparable levels of the trajectories optimized with individual objectives.
\end{abstract}


\section{Introduction}
\label{sec:introduction}
\begin{figure}
    \centering%
    \revisionchange{
    \begin{tikzpicture}
        \node[inner sep=0] (image) at (0,0) {\includegraphics[width=.8\columnwidth]{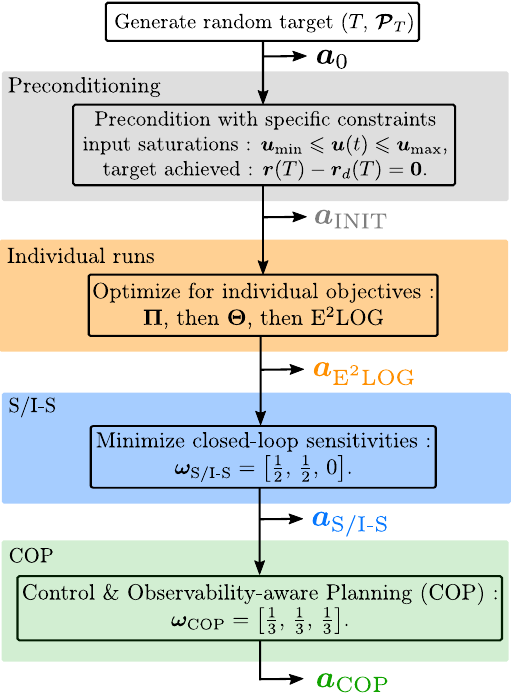}};
        \draw[red,ultra thick] (image.south east) -- (image.north west);
        \draw[red,ultra thick] (image.north east) -- (image.south west);
        \draw[red,ultra thick] (image.south west) rectangle (image.north east);
    \end{tikzpicture}
    \\   }{}     
    \includegraphics[width=.8\columnwidth]{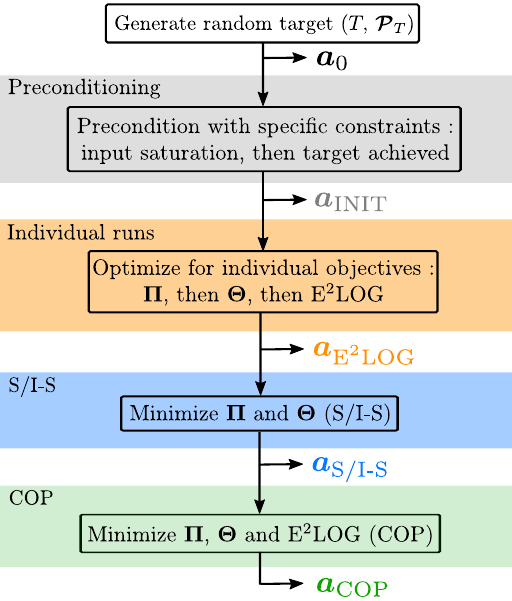}
    \vspace{4pt}
    \caption{Graphic overview of the multi-step \glsfirst{soop} solved with \glsfirst{cop}, $\bm{a}_i$ being the resulting trajectory coefficients after each operation. \revisionchange{The influence of each operation on the trajectory can be seen in \figref{example}.}{}}%
    \label{fig:title}
\end{figure}%

\subsection{Motivation \& Related Work}
\label{sec:related_work}

Motion planning algorithms are a crucial component of autonomous task execution for an \gls{uav} or robots in general. \revisionchange{A simple solution is way-point following through linear control methods. Off-the-shelf auto-pilot software, e.g., PX4 or ArduPilot, offers this method. Such an approach makes motion planning simple as only a few points in space need to be defined. This trivial method comes at the expense of execution time and only allows for conservative motions, often interrupted by full stops.}{}

\subsubsection{Trajectory Generation}
\revisionchange{Another}{A common} approach is to generate a series of\revisionchange{such}{} way-points in discrete time intervals - a so-called \emph{trajectory}. These trajectories are often 4D flat outputs (3D position and yaw orientation) of the \gls{uav} with their subsequent derivatives. These generation methods often let the \gls{uav} move from point A to point B with additional goals and constraints, e.g., reducing flight time or energy consumption. Examples of such generation methods based on continuous-space models \revisionchange{of the \gls{uav}}{}can be found in \cite{berscheid_trajectory_2021,loew_prompt_2021,mellinger_minimum_2011,morbidi_energy_2016}, acting on different levels of abstraction and even taking probabilities into account. \cite{chen_planning_2019,shi_rrt_2020,oleynikova_replanning_2016,candido_pomdp_2010} create a discrete map of the task-space, a so-called sampling-based method, to apply search algorithms to find a path from A to B. With the increase of computational power, learning-based approaches attracted attention in recent years. \cite{liu_learning_2021,schperberg_2021_saber,xi_2020_reinforcement} use machine learning for trajectory generation with reduced calculation times.

\revisionchange{All previously mentioned approaches do not take robustness against model and state uncertainties into account. The result might be non-informative trajectories for state/parameter estimation or trajectories whose tracking by the chosen controller is poorly robust against uncertainties in the model parameters.}{All previously mentioned approaches do not consider model and state uncertainties to ensure robust trajectory tracking or accurate state/parameter estimation.}

\subsubsection{Estimation-aware Trajectories}
State estimation with proper system modeling~\cite{boehm_filter_2021} relies on the design of \emph{sufficiently informative} system input for accurate and fast estimate convergence. The works in \cite{hausman_observability-aware_2017,preiss_simultaneous_2018,boehm_observability_2020,wilson_optimization_2015,ponda_trajectory_2009,berg_lqg-mp_2011,ansari_sensitivity_2016} show that taking the estimation or parametric uncertainty into account drastically improves system parameter estimation results while allowing task execution. In this work we will use the \gls{e2log} \cite{preiss_simultaneous_2018}. These \emph{informative} trajectories might show poor robustness against uncertainties in the robot model for the tracking controller that executes them~\cite{lee_tracking_2010,kamel_mpc_2016}. 

\subsubsection{Control-aware Trajectories}
To address this issue, the notion of \emph{Closed-Loop \gls{sis}} has been recently introduced in~\cite{brault_robust_2021,robuffo_sensitivity_2018} as a suitable metric to be optimized. Minimizing the norm of the \gls{sis} generates a trajectory whose tracking results are minimally sensitive to model uncertainties of the robot states and inputs. This is important as it increases the robustness and accuracy of the trajectory tracking and improves the repeatability of the control inputs when the model parameters vary. The \gls{sis}, however, needs knowledge of the actual robot state and nominal values of the model parameters, which may not be directly available but can be provided by state estimation.

\subsubsection{Link \& Combination}
Therefore, a link between observability-related metrics and \gls{sis} metrics exists, since the evaluation of \gls{sis} needs a good knowledge of states and parameters, and the tracking of maximally observable trajectories benefits from increased robustness in the trajectory execution. That fact motivates our study on how to combine both methods in a unified trajectory optimization problem taking into account that these two objectives can conflict among themselves, as seen in \secref{results}.

\subsection{Contributions}
\label{sec:contributions}

This paper proposes Control \& Observability-aware Planning (COP) as a way to balance two \emph{possibly contradicting} optimization problems, namely $(i)$ generating trajectories whose execution is \emph{minimally sensitive} to model uncertainties and $(ii)$ generating trajectories that can be \emph{sufficiently informative} for accurate state estimation. 
We present a method to address these problems by leveraging previous contributions to the topics of observability-aware and minimally-sensitive trajectory planning, combining them in the formulation and resolution of a \gls{soop} based on the Augmented Weighted Tchebycheff method. In a case study, considering the state estimation and robust trajectory tracking for a 3D quadrotor \gls{uav}, we discuss the statistical results of a realistic simulation campaign that shows the potential of the proposed contribution.


\section{Preliminaries}
\label{sec:preliminaries}
\revisionchange{This section serves the purpose of explaining the changes in the quadrotor model as well as the controller in the transition to 3D, compared to \cite{brault_robust_2021}.}{}

\subsection{Quadrotor model}
\label{sec:model}
Let us consider a frame $\mathcal{M}$ as the quadrotor body frame, attached to its \gls{com}, with its z-axis $\bm{z}_M$ aligned with the thrust of the four rotors. The state vector $\vx$ of this system consists of its linear position $\bm{r} = (x,\, y,\, z)$ and velocity $\bm{v} = (v_x,\, v_y,\, v_z)$, both expressed in the world frame $\mathcal{W}$. It also includes the body orientation expressed through the unit length quaternion $\bm{q} = (q_w,\, q_x,\, q_y,\, q_z)$ (Tait–Bryan angle definition with yaw first (312-sequence)) as well as its angular velocity $\bm{\omega} = (\omega_x,\, \omega_y,\, \omega_z)$, expressed in the body frame $\mathcal{M}$, therefore \revisionchange{$\vx = \left[\bm{r}\ \bm{v} \ \bm{q} \ \bm{\omega}\right]^{\transpose}\in \mathbb{R}^{13}$}{$\vx = \left[\bm{r}_{}^{\transpose}, \bm{v}_{}^{\transpose}, \bm{q}_{}^{\transpose}, \bm{\omega}_{}^{\transpose}\right]^{\transpose}\in \mathbb{R}^{13}$}.


As the quadrotor's orientation is expressed by the \revisionchange{}{Hamiltonian} quaternion $\bm{q}$, we recall that $\bm{z}_M = \vR(\bm{q})\bm{z}_W \revisionchange{= \mathcal{R}(\bm{q}, \, \bm{z}_W )}{} = \bm{q}^{-1} \otimes \bm{z}_W  \otimes \bm{q}$, where $\otimes$ is the quaternion product. $\vR(\bm{q})$ is the rotation matrix as a function of the quaternion $\vq$. \revisionchange{$\mathcal{R}$ is the quaternion linear rotation operator, which rotates a vector with a unit quaternion, without modifying its magnitude. Analogously, $\bm{x}_M = \mathcal{R}(\bm{q}, \, \bm{x}_W)$  and $\bm{y}_M = \mathcal{R}(\bm{q}, \, \bm{y}_W)$.}{}

It is possible to link the squared rotor speeds of the quadrotor (control inputs) $\vu = \left[\omega_1^2\ \omega_2^2 \ \omega_3^2 \ \omega_4^2\right]^{\transpose}$ to \revisionchange{$(f,\, \bm{\tau})$}{$[f,\, \bm{\tau}_{}^{\transpose}]_{}^{\transpose}$}, the total effective thrust and torque. \revisionchange{These are related by 
\begin{equation}
\begin{bmatrix} 
f \\[0.5em]
\tau_x\\[0.5em]
\tau_y\\[0.5em]
\tau_z\\[0.5em]
\end{bmatrix} 
= 
k_f\begin{bmatrix}
1 & 1 & 1 & 1\\[0.5em]
0 & \ell & 0 & -\ell\\[0.5em]
-\ell & 0 & \ell & 0\\[0.5em]
k_m & -k_m & k_m & -k_m
\end{bmatrix} 
\begin{bmatrix}
\omega_1^2 \\[0.5em]
\omega_2^2 \\[0.5em]
\omega_3^2\\[0.5em]
\omega_4^2\\[0.5em]
\end{bmatrix}
= 
\bm{S}\cdot \bm{u},
\label{eq:allocmatrix}
\end{equation}
where $\bm{S}$ is the allocation matrix which includes the rotor thrust force coefficient $k_f$, the drag moment coefficient $k_m$, and the arm length $\ell$ from the center of mass to each motor/rotor group. Matrix $\bm{S}$ is invertible: for a given $(f,\, \bm{\tau})$, the speed of each rotor can be obtained from \equref{allocmatrix}.}{These are related by the allocation matrix $\bm{S}$, $[f,\, \bm{\tau}_{}^{\transpose}]_{}^{\transpose}=\bm{S}\vu$ (e.g., \cite{mahony_multirotor_2012} Eq. (8)), which includes the rotor thrust force coefficient $k_f$, the drag moment coefficient $k_m$, and the arm length $\ell$ from the center of mass to each motor/rotor group}. 

With these definitions, \revisionchange{}{compared to \cite{brault_robust_2021},} the quadrotor dynamical model is
\begin{equation}
\dot{\vx} = \left\lbrace\begin{aligned}
\bm{\dot{r}}_W &= \bm{v}_W \\
\bm{\dot{v}}_W &= -g\bm{z}_W + \tfrac{f}{m} \bm{z}_M \\
\bm{\dot{q}} &= \tfrac{1}{2} \bm{q}\otimes \omega_M\\
\bm{\dot{\omega}}_M &= \bm{J}^{-1}(\bm{\tau} - \bm{\omega}_M\times (\bm{J}\bm{\omega}_M))
\end{aligned}\right.
\label{eq:quadrotordq}
\end{equation}
where $m$ and $\bm{J}$ are the quadrotor's mass and its inertia matrix, respectively. $g$ is the Earth’s gravitational pull. \revisionchange{\equref{allocmatrix} implies that the dynamics are not only affected by the parameters $m$ and $\bm{J}$, but also by the parameters $k_f$, $k_m$ and $\ell$.}{The allocation matrix $\bm{S}$ implies that the dynamics are also effected by the parameters $k_f$, $k_m$ and $\ell$.} The quantities $k_f$ and $k_m$ are aerodynamic parameters that depend on the rotors characteristics and the aerodynamic interaction with the environment (e.g., presence of wind, ground effects). Therefore, an accurate value for these parameters can be hard to obtain, and we thus consider $\vp = \left[ k_f \ k_m \right]^T \in \mathbb{R}^2$ as the uncertain parameters of the dynamical model. 

\subsection{Tracking controller}
\label{sec:uav_control}

With the dynamic model detailed in the previous subsection, we can now present the controller that has been used in this work. The chosen control task is to let the system output $\bm{y}(\vx) = \left[ x\ y\ z\ \varphi \right]^{\transpose} \in \mathbb{R}^{4}$ track a desired motion $\bm{r}_d(\bm{a},\ t) \in \mathbb{R}^4$, where $\varphi$ is the yaw (or heading) angle of the quadrotor\revisionchange{, and can easily be linked to the quaternion $\bm{q}$}{}. \revisionchange{In \cite{robuffo_sensitivity_2018, brault_robust_2021}, a DFL (Dynamic Feedback Linearization) controller with integral term was used as tracking controller. The DFL law is able to perfectly track any desired motion (in absence of input constraints and uncertainties in the states and parameters). However, this type of control is rarely used in practice because of its complexity (introduction of additional states with complex dynamics) and poor robustness against uncertainties.}{The DFL (Dynamic Feedback Linearization) controller with an integral term used in previous works \cite{robuffo_sensitivity_2018, brault_robust_2021} is not robust against parameter uncertainties and is not capable of considering input constraints.} Therefore, we chose to use a different controller, namely the so-called Lee controller~\cite{lee_tracking_2010}, which performs slightly worse in an ideal case compared to the DFL, but it is much less complex to implement and tune\revisionchange{ (indeed, the use of the Lee controller and variants is widespread in the community)}{}.

Although we implemented the same structure of the controller as in \cite{lee_tracking_2010}, we added some minor changes in order to match the dynamics of the quadrotor in 3D (especially the use of the quaternion $\vq$). In particular 
we consider attitude and angular velocity errors defined as
\revisionchange{
\begin{equation}
\bm{e}_{\bm{q}} = \dfrac{1}{2}\left(\bm{R}_{d}^{\transpose}\bm{R}(\bm{q}) - \bm{R}(\bm{q})^{\transpose}\bm{R}_{d}\right)^{\vee}
    \label{eq:attitudeerror}
\end{equation}
where $\bm{R}(\bm{q})$ is the rotation matrix function of $\bm{q}$, and
\begin{equation}
\bm{e}_{\bm{\omega}} = \bm{w}.
    \label{eq:angvelerror}
\end{equation}
}{
\begin{equation}
    \bm{e}_{\bm{q}} = \dfrac{1}{2}\left(\bm{R}_{d}^{\transpose}\bm{R}(\bm{q}) - \bm{R}(\bm{q})^{\transpose}\bm{R}_{d}\right)^{\vee} \text{ and } \bm{e}_{\bm{\omega}} = \bm{w}. \label{eq:controlerror}
\end{equation}
}
The resulting control inputs are then
\begin{equation}
f = \left(-\bm{k}_r \bm{e_r} - \bm{k}_v \bm{e_v} - \bm{k}_i \bm{\xi} + m(g\bm{z}_{w} + \ddot{\bm{r}}_d) \right) \cdot \bm{R}(\bm{q})\bm{z}_{w},
    \label{eq:desiredthrust}
\end{equation}
\begin{equation}
\bm{\tau} = -\bm{k}_q \bm{e_q} - \bm{k}_{\omega} \bm{e_{\omega}},
    \label{eq:desiredtorque}
\end{equation}
where $\bm{e}_r$ and $\bm{e}_v$ are the position and velocity errors, and $\bm{\xi} = \left[ \xi_x\ \xi_y \ \xi_z \right]^{\transpose}$ is the position integrator, and $\bm{k}_r$, $\bm{k}_v$, $\bm{k}_i$, $\bm{k}_q$, $\bm{k}_{\omega}$ are suitable control gains. We then compute $\bm{u}$ via the inverse of the allocation matrix\revisionchange{}{, $\vu=\bm{S}_{}^{-1}[f,\, \bm{\tau}_{}^{\transpose}]_{}^{\transpose}$.} 
\revisionchange{
\begin{equation}
\bm{u} = \bm{S}_c^{-1} \cdot \begin{pmatrix}
f \\
\bm{\tau}
\end{pmatrix}.
    \label{eq:ftautou}
\end{equation}
}{}


\subsection{Curve representation}
\label{sec:curve}

The controller is designed to let the quadrotor follow a reference trajectory $\bm{r}_d(\bm{a},\ t)$, where $\bm{a}$ is the parameter vector for the chosen class of curve. In \cite{robuffo_sensitivity_2018} `plain polynomials' were used, with the drawback of introducing possible numerical instability during the optimization. 
\revisionchange{Thereby in~\cite{brault_robust_2021} we switched to the use of Bézier curves, since, in this case, adjusting a control point in its admissible space during the optimizations is in general quite stable from a numerical point of view.}{Due to this reason, we switched in \cite{brault_robust_2021} to the use of Bézier curve representation, as they are more stable from a numerical point of view.} 

\revisionchange{In this work, in order to gain an even more intuitive and flexible trajectory representation, we adopted piecewise Bézier curves, instead of only a single one of high degree as in~\cite{brault_robust_2021}. In this new implementation, the parameters are not the control points, but way-points along this curve that set the limit conditions at the beginning and the end of the whole trajectory, as well as between two pieces of the trajectory. The trajectory passes through the way-points with desired velocities and accelerations (and even higher derivatives if needed). We now give some details about this representation, and show how to translate the limit conditions into Bézier curves.}{This work goes one step further by implementing piecewise Beziér curves for the trajectory representation to avoid the use of a single high degree Beziér curve as in~\cite{brault_robust_2021}. An aditional abstraction happens on the parameter vector $\bm{a}$, which now contains way-points (with velocity, acceleration, and even higher order constraints) instead of the control points themselves. These way-points define the curve at the start, end, and in between curve pieces.}

Let $\mathcal{B}_{i\in \llbracket 1, \, n\rrbracket}$, with $n\geqslant 1$ number of pieces, be the Bézier curve of degree $d \geqslant 2$ shaping the trajectory\revisionchange{, then $\mathcal{C}^{d-1}$ continuity is assured between each piece}{ ($\mathcal{C}^{d-1}$ continuity)}. In total, there are $n+1$ way-points $(t_i,\, \bm{\mathcal{P}}_i)$, where $t_i$ is the time associated to the point $\bm{\mathcal{P}}_i \in \mathbb{R}^{n_{\text{dim}} \times n_{\text{jc}}}$, with $n_{\text{dim}}$ the number of dimensions of the trajectory (\revisionchange{for instance, for a quadrotor}{e.g.}, $x$, $y$, $z$, and \revisionchange{its}{}yaw angle), and $n_{\text{jc}}$ the number of joining conditions (e.g., $n_{\text{jc}} = 1$ for position only, which means every Bézier curve piece is a straight line segment). The degree $d$ of the Bézier curve, the number of joining conditions $n_{\text{jc}}$ and the number of control points $n_{c}$ are linked by \revisionchange{}{$d = 2n_{\text{jc}} - 1 = n_{c} - 1$.}%
\revisionchange{
\begin{equation}
d = 2n_{\text{jc}} - 1 = n_a - 1.
    \label{eq:beziercurvenumbers}
\end{equation}
}{}
\revisionchange{With the conditions mentioned above, it is possible to use the derivative of the Bézier curve to construct all the joining conditions. Then, one can define the linear system of equations which allows to obtain the control points of each piece from its limit conditions. The solution of this system outputs the trajectory as a tensor containing the way-points, of dimension $(n+1) \times n_{\text{jc}} \times n_{\text{dim}}$ and also the vector $t_{i\in \llbracket 1,\, n+1 \rrbracket}$ containing the time of each way-point.}{With these conditions and the way-points as constraints, one can formulate a linear system of equations that solves for the control points of each Bezièr curve piece.}


\section{Control \& Observability-aware Planning}
\label{sec:combined}
\subsection{Objectives for Trajectory Optimization}
A trajectory $\bm{r}_d(\bm{a},t)$ parameterized by the coefficient vector $\bm{a}$, \secref{curve}, can be optimized for different goals by changing its shape. We represent this goal by the so-called utility function $U(\bm{a})$ which, in our case, is a scalar cost to be minimized subject to constraints
\begin{equation}
    \label{eq:fmincon}
    \begin{aligned}
        & \underset{\bm{a}}{\text{minimize}}
        & & U(\bm{a}), \\
        & \text{subject to}
        & & \bm{a} \in \mathcal{A},\\
        &&& \text{equ. \& inequ. constraints},
    \end{aligned}
\end{equation}
with $\mathcal{A}$ as the feasible set for the parameter vector.

\subsubsection{State Sensitivity Metric}
The state sensitivity metric as minimization objective was introduced in \cite{giordano_closed-loop_2018} and is based on a generic robot model $\dot{\vx} = \vf(\vx,\,\vu,\,\vp)$, where $\vx\in\mathbb{R}^{n_x}$ is the state vector, $\vu\in\mathbb{R}^{n_u}$ is the control input vector, and $\vp\in\mathbb{R}^{n_p}$ the vector of \emph{system} parameters (which are assumed uncertain). This is combined with a tracking control law $\dot{\bm{\xi}} =\vg(\bm{\xi},\,\vx,\,\bm{r}_d(\va,\,t),\,\vp_c)$ and $\vu = \vc(\bm{\xi},\,\vx,\,\bm{r}_d(\va,\,t),\,\vp_c)$, where $\bm{\xi}\in\mathbb{R}^{n_\xi}$ are the internal controller states (e.g., an integrator), $\vp_c$ a nominal value for the parameters $\vp$, and $\bm{r}_d(\va,\,t)$ a desired trajectory\revisionchange{ to be tracked parameterized by vector $\va$ and time $t$}{}.
The state sensitivity matrix for the \emph{closed-loop system} (i.e., considering both the robot and chosen control) is defined as
\begin{equation}
    \label{eq:Pi}
    \left.
    \vPi(t) = \dfrac{\partial  \vx(t)}{\partial \vp}
    \right|_{\vp = \vp_c}
    \in \mathbb{R}^{n_x \times n_p}
\end{equation}
representing the variation of the states $\vx$ in relation to variations in the parameter vector $\vp$, evaluated on the \textit{nominal} value $\vp_c$. We refer the reader to \cite{giordano_closed-loop_2018} for further details. 

The integral of the matrix norm of the state sensitivity over the whole trajectory (duration $T$) can be used to reduce the influence of parameter uncertainty on the states.
\begin{equation}
    \label{eq:u_pi}
    U(\bm{a}) = F_{\Pi}^{}(\bm{a}) = \int_0^T||\bm{\Pi}(t)||\,\mathrm{dt}
\end{equation}

\subsubsection{Input Sensitivity Metric}
As natural evolution, \cite{brault_robust_2021} added to the state sensitivity metric the so-called input sensitivity metric which is defined as
\begin{equation}
    \label{eq:Theta}
    \left.
    \bm{\Theta}(t) = \dfrac{\partial \vu(t)}{\partial \vp}
    \right|_{\vp = \vp_c}
    \in \mathbb{R}^{n_u \times n_p}
\end{equation}
and, similarly to \equref{Pi}, maps how variations of the parameter vector $\vp$ result in variations of the control inputs. The integral of its matrix norm, again, gives us the objective function \equref{u_theta} which can be used to reshape the trajectory to be less sensitive in its control inputs against parameter uncertainties. Details on the derivations can be found in \cite{brault_robust_2021}.
\begin{equation}
    \label{eq:u_theta}
    U(\bm{a}) = F_{\Theta}^{}(\bm{a}) = \int_0^T||\bm{\Theta}(t)||\,\mathrm{dt}
\end{equation}

\revisionchange{
In general, it is not possible to compute $\bm{\Pi}(t)$ in closed-form. However, it is possible to obtain a closed-form expression for the dynamics of $\bm{\Pi}(t)$ from which any other needed quantity can be computed. Therefore, the evolution of ${\vPi}(t)$ over a time interval $T$ of interest can be obtained by numerically integrating over time, \cite{giordano_closed-loop_2018,brault_robust_2021}. Note that the evaluation of ${\vPi}(t)$ and $\bm{\Theta}(t)$ is done by leveraging both the robot's model $\vf(\cdot)$ and the control law $(\vg(\cdot),\,\vc(\cdot))$, so that any strength/weakness of the chosen control action is correctly taken into account. This is why these quantities are also called \emph{closed-loop} state and input sensitivities, since they are evaluated on the full closed-loop system. Brought together for trajectory planning, both these objectives provide the ability to generate \textit{control-aware} trajectories, see \secref{sis}. That being said, one of the main hypothesis in \cite{giordano_closed-loop_2018,brault_robust_2021} is to consider that the state of the actual system is fully known during the tracking of the desired task. Thereby, ensuring that the state is best known is one of the key remaining aspects that need to be solved for this kind of trajectory planning.
}{
Note that it is not possible to compute $\bm{\Pi}(t)$ in closed-form. However, it is possible to obtain a closed-form expression of its dynamics, and other quantities, e.g., $\bm{\Theta}(t)$, can be derived from it. The evolution of ${\vPi}(t)$ (consequently $\bm{\Theta}(t)$ too) over a time interval $T$ of interest then can be obtained by numerically integrating over time~\cite{giordano_closed-loop_2018,brault_robust_2021}. One of the main hypotheses/key aspects in \cite{giordano_closed-loop_2018,brault_robust_2021} is to consider that the whole state of the system is known during the tracking of the desired trajectory.
}

\subsubsection{Observability Metric}
Online state estimation is one way of making the state $\vx$ and parameters $\vp$ available at run-time\revisionchange{}{~\cite{weiss_versatile_2012,hausman_self-calibrating_2016,boehm_filter_2021}}. \revisionchange{This is possible through the system model $\vf(\cdot)$ and the comparison of sensor measurements with system state-based measurement models $\vh(\vx,\vu,\vp)\in\mathbb{R}^{n_h}$,  \cite{weiss_versatile_2012,hausman_self-calibrating_2016,boehm_filter_2021}.}{}

How well such estimates perform depends on the accuracy of the system model and sensors used, but control inputs given to the robot are equally important. The \gls{e2log} \cite{preiss_simultaneous_2018,hausman_observability-aware_2017} works on the idea of the quality of observability, proposed in \cite{hermann_nonlinear_1977, krener_measures_2009}, which evaluates it over a whole trajectory $\bm{r}_d(\va,\,t)$ with a duration $T$. It uses a $n$-th order Taylor expansion to approximate the Jacobian matrix \revisionchange{$\vK_{\va,t_0}$ of \revisionchange{$\vh(\cdot)$}{the measurement model $\vh(\vx,\vu,\vp)\in\mathbb{R}^{n_h}$} at time $t_0$. This approximation}{which} models the sensitivity of the measurements with respect to the control inputs, the state and its changes on a small time horizon $H$\revisionchange{ of the trajectory}{} ($\widetilde{\vW}_{t_0,H}(\va)$ in \cite{preiss_simultaneous_2018}).
Summing all these quality measure segments along the trajectory $\bm{r}_d(\va,\,t)$ gives us\revisionchange{ the \gls{e2log}}{}
\begin{equation}
    \label{eq:e2log}
    \widetilde{\vW}_{\cO}(\va) = \sum_{k=0}^{N} \widetilde{\vW}_{k\Delta t,\Delta t}(\va)  \in \mathbb{R}^{n_x \times n_x}
\end{equation}
with $\Delta t = \tfrac{T}{N}$ and $N$ the number of trajectory segments.

The objective is to improve the least sensitive state (or combination of states) through the smallest eigenvalue of $\widetilde{\vW}_{\cO}(\va)$. Adding a minus to the smallest eigenvalue makes it usable in a minimization problem as objective function
\begin{equation}
    \label{eq:u_e2log}
    U(\bm{a}) = F_{\text{E$^2$LOG}}^{}(\bm{a}) = -\sigma_{\min}(\widetilde{\vW}_{\cO}(\va)).
\end{equation}

The result is a trajectory with optimum observability properties, improving the states' convergence by lowering the uncertainty and increasing the overall accuracy.

The estimator in [12] is the base for the derivation of the observability-aware trajectory optimization in our quadrotor \gls{uav} case study, \secref{results}.

\subsection{Multi-Objective Optimization Problem}
\label{sec:moop}
The problem presented in this work is an example of a \gls{moop} trying to optimize for different objectives with constraints.

\subsubsection{Pareto Optimality}
Ideally, one would try to find the non-dominated set in the entire feasible set $\mathcal{A}$ for the parameter vector $\bm{a}$, a so-called Pareto optimal set \revisionchange{}{or Pareto front} \cite{steuer_interactive_1983,wierzbicki_multiple_1995,marler_survey_2004,dachert_adaptive_2010,emmerich_tutorial_2018,holzmann_solving_2018}. The term \emph{non-dominated} or \emph{optimal} means that the current set does not improve one objective while worsening another. We compute only one point in the Pareto optimal set due to the complexity of the objective functions and the resulting computation times, see \secref{times}.

In \cite{brault_robust_2021}, \gls{lsp} was chosen as a method because it achieved good results in balancing the \gls{sis} of a 2D quadrotor. \gls{lsp} combines objective functions to a single cost through a linear weighting of each objective\revisionchange{, to allow for a single-objective constraint optimization}{}. This is only a feasible option if the Pareto front\revisionchange{ (set of all optimal solutions)}{} is convex. 
\revisionchange{\gls{lsp} comes with its drawbacks as well, often caused by the individual objective functions not building a convex front.}{} To be more specific, in the case of concave Pareto fronts, \gls{lsp} tends to converge towards extrema solutions, an optimal solution for only one of the objectives. Evenly distributed weights do not produce an evenly distributed representation of the Pareto optimal set. The addition of the observability-awareness through the \gls{e2log} as a third objective is the natural evolution of the approach\revisionchange{ to improve estimation performance}{}, however, non-convexity (concavity) is possible with this addition. 

\subsubsection{Augmented Weighted Tchebycheff Method}
\label{sec:tcheb}

\revisionchange{To solve this issue, we use the Augmented Weighted Tchebycheff method to balance all three objectives or any combination, compared to \cite{brault_robust_2021}.}{}
\revisionchange{This is done}{To solve this issue, we balance the objectives} through their distance between the objective value $F_{i}^{}(\bm{a})$ and an aspiration point $F_{i}^{O}$\revisionchange{}{, compared to \cite{brault_robust_2021}}. \revisionchange{In this work, this}{This} aspiration point $F_{i}^{O}$ is the individual objective's minimal value from \equref{fmincon} -\revisionchange{ then}{} called \emph{utopia point}. Another important point in the Pareto set is the so-called \emph{nadir point} $F_{i}^{N} = \underset{1 \leq j \leq k}\max\{ F_{i}^{}(\bm{a}_{j}^{O}) \}$. It is the largest cost of all objectives with respect to the $j$-th utopia point.\revisionchange{ The utility function is defined as}{}
\begin{equation} 
    \label{eq:cop_problem}
    U(\bm{a}) = \underset{i}\max \{ \lambda_{i}^{} |F_{i}^{}(\bm{a})-F_{i}^{O}| \} + \rho \sum_{j=1}^{k} |F_{j}^{}(\bm{a})-F_{j}^{O}|
\end{equation}
\revisionchange{ where}{The utility function in \equref{cop_problem} includes} $\vF(\bm{a}) = \{ F_{\Pi}^{}(\bm{a}), F_{\Theta}^{}(\bm{a}), F_{\text{E$^2$LOG}}^{}(\bm{a}) \}$, and the scale of each objective $\lambda_{i}^{} = \frac{w_{i}^{}}{|F_{i}^{N}-F_{i}^{O}|}$. Every point on a Pareto front is a minimum of the Tchebycheff function for some $\lambda_{i}^{}$ (convex or non-convex) and achieves Pareto optimality of the solution. The weight $w_{i}^{}$ is the user defined preference of one objective, which selects one solution from the possible set of Pareto optimal solutions (bias), $\sum_{i=1}^{k} w_{i}^{} = 1$. $|F_{i}^{N}-F_{i}^{O}|$ in the denominator of $\lambda_{i}^{}$ normalizes the Tchebycheff function to the interval $[0,1]$. According to \cite{steuer_interactive_1983}, $\rho$ values should be selected between 0.0001 and 0.01. We refer the reader to \cite{wierzbicki_multiple_1995,marler_survey_2004,emmerich_tutorial_2018} for further reading. As can be seen, this method\revisionchange{ also}{} reduces the \gls{moop} into a \gls{soop}\revisionchange{ with constraints}{}. We will discuss how to use \equref{cop_problem} in \secref{sis} and \secref{cop}.

\subsection{Implementation}
\label{sec:implementation}

\gls{cop} uses a multi-step approach to the trajectory optimization\revisionchange{}{, see \figref{title}}, as the utopia point $F_{i}^{O}$ and nadir point $F_{i}^{N}$ of each objective $F_{i}$ need to be known before combining objectives\revisionchange{, \secref{tcheb}}{}. The framework is implemented in Python and uses a local derivative-free optimization, namely \gls{cobyla} of the open-source library for nonlinear optimization (NLOPT). The numerical integration method dopri5 of SciPy enables us to calculate the individual costs (\equref{u_pi}, \equref{u_theta}, and \equref{u_e2log}) during each iteration of the optimization.\revisionchange{ \figref{title} gives a graphical overview of the \gls{cop} method, and one can see that \gls{cop} returns all possible optimized trajectories.}{}

\revisionchange{
\begin{figure}[t]
    \centering%
    \vspace{1em}
    \revisionchange{
    \begin{tikzpicture}
        \node[inner sep=0] (image) at (0,0) {\includegraphics[width=0.8\columnwidth]{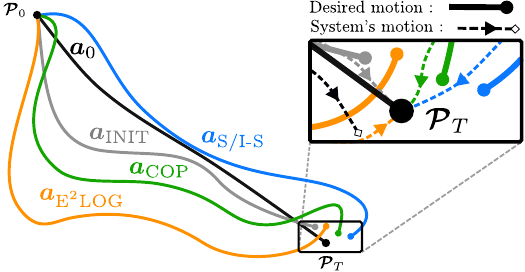}};
        \draw[red,ultra thick] (image.south east) -- (image.north west);
        \draw[red,ultra thick] (image.north east) -- (image.south west);
        \draw[red,ultra thick] (image.south west) rectangle (image.north east);
    \end{tikzpicture}
    \\   }{}        
    \vspace{4pt}
    \caption{
    \revisionchange{Example of \gls{cop} for the target $\bm{P}_T$ : $\bm{a}_0$ contains the initial and final way-points, $\bm{P}_0$ and $\bm{P}_T$, and the free decision variables. $\bm{a}_{\text{INIT}}$ is the preconditioned trajectory, and $\bm{a}_{\text{E$^2$LOG}}$, $\bm{a}_{\text{S/I-S}}$, and $\bm{a}_{\text{COP}}$ are the output trajectories of the different optimization runs. The zoomed portion near the final way-point $\bm{P}_T$ shows that only $\bm{a}_0$ goes to the target, resulting in a motion offset to the desired position. For all other trajectories, the desired motion ensures the real system reaches the target in the nominal case.}{}}%
    \label{fig:example}
\end{figure}
}{}

\subsubsection{Preconditioning}
\label{sec:preconditioning}
\revisionchange{It is the first and necessary step in the trajectory generation, to ensure that the initial trajectories are dynamically feasible and let the system reach the final target accurately.}{This important step ensures that the initial trajectories are dynamically feasible and let the system reach the final target accurately.} The preconditioning starts with a random trajectory with initial way-point $\bm{\mathcal{P}}_0$ (for a 3D quadrotor \gls{uav} position and yaw orientation), target way-point $\bm{\mathcal{P}}_T$, duration $T$, and $n$ number of Bézier curve pieces supplied. All way-points between the initial and target represent the decision variables $\bm{a}_i$ of the optimization, and can be chosen freely within the admissible set $\mathcal{A}$.
Then, dynamical constraints are applied to the trajectory through a short optimization (e.g., min. and max. rotor speeds). \revisionchange{Afterwards, we precondition the trajectory to take the controller tracking imperfections into account, resulting in a trajectory in which the last way-point may differ from the real target (thus, different from $\bm{\mathcal{P}}_T$). Thereby, the tracking of this new trajectory ensures that the system reaches the final target accurately, only with the \textit{nominal} value $\vp = \vp_c$, thus, without uncertainties in the model, depicted in \figref{example}.}{Afterwards, we precondition the trajectory to take the controller tracking imperfections into account, ensuring that the system reaches the final target accurately with the \textit{nominal} value $\vp = \vp_c$, thus, without uncertainties in the model.} The vector $\bm{a}_{\text{INIT}}$ parameterizes the shape of the\revisionchange{ resulting}{} initial trajectory.\revisionchange{ This trajectory serves as a baseline\revisionchange{ in the latter evaluation}{} in \secref{results}.}{}

\subsubsection{Optimizing Individual Objectives}
\label{sec:individual}
As mentioned before, to allow the combination of multiple objectives, one needs to compute $F_{i}^{O}$ and $F_{i}^{N}$ of each objective first. We do the minimization of each objective $F_{\Pi}^{}$, $F_{\Theta}^{}$, and  $F_{\text{E$^2$LOG}}^{}$, defined in \equref{u_pi}, \equref{u_theta}, and \equref{u_e2log} respectively, along with \equref{fmincon} to get those needed extrema points. The results are the utopia point and coefficient vector of each objective, $F_{\Pi}^{O}$/$\bm{a}_{\Pi}^{O}$, $F_{\Theta}^{O}$/$\bm{a}_{\Theta}^{O}$, and $F_{\text{E$^2$LOG}}^{O}$/$\bm{a}_{\text{E$^2$LOG}}^{O}$, respectively. These coefficient vectors are used to calculate the nadir points of the objectives $F_{\Pi}^{N}$, $F_{\Theta}^{N}$, and $F_{\text{E$^2$LOG}}^{N}$. 

\subsubsection{State and Input Sensitivities Optimization}
\label{sec:sis}
A first example for \equref{cop_problem} is the computation of the \gls{sis}. The balancing of state and input sensitivities similar to \cite{brault_robust_2021} is possible by setting $\bm{w}_{\text{\gls{sis}}}^{} = \left[\tfrac{1}{2},\tfrac{1}{2},0\right]$ and using the utopia and nadir points of $F_{\Pi}^{}$ and $F_{\Theta}^{}$ for normalization. The rationale behind $\bm{w}_{\text{\gls{sis}}}^{}$ is as follows: $(i)$ \revisionchange{the first objective, }{}\equref{u_pi}\revisionchange{,}{} aims at minimizing the state sensitivity over the trajectory, so that the tracking accuracy of $\bm{r}_d(\va,\,t)$ is made most insensitive to uncertainties in the model parameters; $(ii)$ \revisionchange{the second objective, }{}\equref{u_theta}\revisionchange{,}{} aims to minimize the input sensitivity during the whole trajectory, in order to obtain control inputs that are most insensitive\revisionchange{ (i.e., with minimal deviations)}{} against variations of the robot parameters. For this optimization we chose $\rho = 0.0001$ in \equref{cop_problem} and the results of this step are $F_{\text{\gls{sis}}}^{O}$/$\bm{a}_{\text{\gls{sis}}}^{O}$. \revisionchange{}{Both these objectives provide the ability to generate \textit{control-aware} trajectories.}

\subsubsection{Control \& Observability-aware Optimization}
\label{sec:cop}
The possible antagonistic nature of the \gls{sis} and \gls{e2log} needs utility functions like \equref{cop_problem} to be balanced successfully. We weight all three objectives equally, $\bm{w}_{\text{\gls{cop}}} = \left[\tfrac{1}{3},\tfrac{1}{3},\tfrac{1}{3}\right]$, in \equref{cop_problem} with $\rho = 0.0001$. This makes the combination of State and Input Sensitivity-based with Observability-aware trajectory planning possible\revisionchange{. The result of this final step are}{ and results in} $F_{\text{\gls{cop}}}^{O}$/$\bm{a}_{\text{\gls{cop}}}^{O}$.

The even distribution of weights\revisionchange{ (taking one point of the Pareto front)}{} might not result in equally improved objectives in practice, due to the possible skewed normalization from the approximation of nadir and utopia points. To ensure we get proper trajectories that reduce both, we apply filtering based on the costs history available from each optimization run - a \emph{posterior} preference $F_{\text{\gls{sis}}}^{} (\bm{a}_{\text{\gls{cop}}}^{O}) < F_{\text{\gls{sis}}}^{} (\bm{a}_{\text{INIT}})$ and $F_{\text{E$^2$LOG}}^{} (\bm{a}_{\text{\gls{cop}}}^{O}) < F_{\text{E$^2$LOG}}^{}(\bm{a}_{\text{INIT}})$.\revisionchange{ \gls{cop} rejects trajectories that do not fulfill these requirements.}{}


\section{Results}
\label{sec:results}
\revisionchange{In this section, we show how the proposed \gls{cop} approach (combination of state and input sensitivities with observability-awareness) can be used to generate trajectories that are less sensitive to changes in the model parameters on the command side, and also favor certain states in the estimation. As a case study, we focus on a 3D quadrotor \gls{uav}'s rotors thrust force and drag moment coefficients $k_f$ and $k_m$.
These parameters were chosen because they have a significant impact on the tracking performance and are hard to estimate.
Furthermore, we only focus on the 3D position $\bm{r}=(x,\,y,\,z)$ of the \gls{uav} as state of interest, without considering the other states in the sensitivity evaluation.}{To show the use of our proposed approach, we conduct a case study focusing on a 3D quadrotor \gls{uav}'s rotors thrust force and drag moment coefficients $k_f$ and $k_m$ ($\bm{p} = \left[ k_f \ k_m \right]^T$). These coefficients have a significant impact on the tracking performance of the control and are hard to estimate.}

\begin{figure}[t]  
    \centering
    \vspace{1mm}
    \includegraphics[width=\columnwidth]{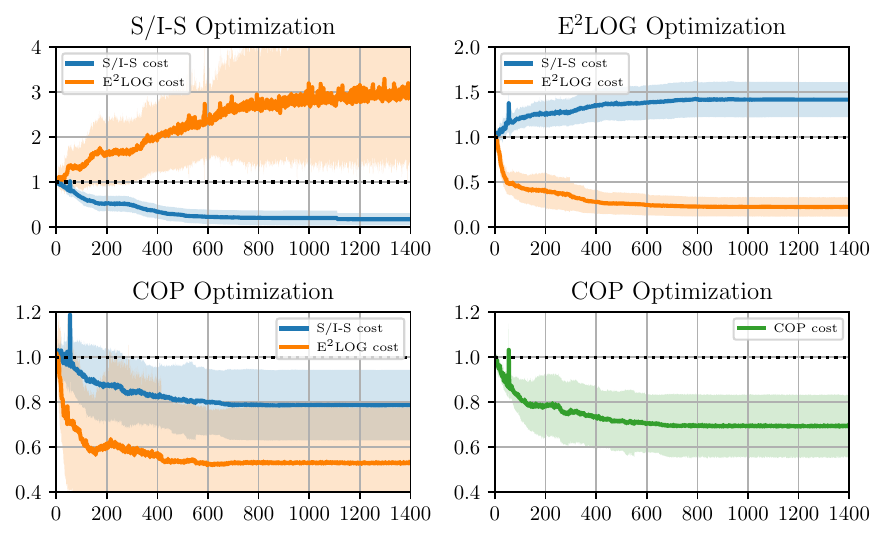}    
    \caption{Overview of the cost function evolution during the optimization depicted by averages and $1\sigma$ standard deviations over 20 optimization runs with different initial trajectories. All cost values are normalized with the initial cost prior to the mean calculation. \gls{sis} cost (blue), \gls{e2log} cost (orange), \gls{cop} cost (green). All optimizations minimize their respective cost function, therefore, a decrease below 1 (dotted line) is an improvement of the respective objective. Note that the inverse of the \gls{e2log} is used.}
    \label{fig:convergence}
    \vspace{-1mm}
\end{figure}

\subsection{Setup \& Evaluation Method}

The system model of \secref{model}, control law of \secref{uav_control}, and optimization of \secref{combined} are implemented in Python. All system parameters are based on the quadrotor Hummingbird model of \cite{boehm_filter_2021}. 
This previous work shows that it is possible to estimate the thrust force and drag moment coefficients, $k_f$ and $k_m$.\revisionchange{ The estimation only needs rotor speed input, position sensor and \gls{imu} measurements, and few \emph{a priori} measurements of the quadrotor.}{}
The method of using simulations allows for better repeatability of experiments and avoids the introduction of other artifacts due to uncertainties of the real system.

In this empirical evaluation we look at four types of trajectories: $(i)$ the initial preconditioned trajectory (INIT); $(ii)$ the \gls{sis} objective optimized one; $(iii)$ the \gls{e2log} objective optimized trajectory; $(iv)$ the new \gls{cop} objective optimized trajectory. The initial one serves as baseline for all other trajectories\revisionchange{ to compare}{}. Each trajectory has a duration of $T = 20s$, 5 Bézier curve pieces, and a random\revisionchange{ly generated}{} target way-point $\bm{P}_T$ in 3D space with $\left(\cU(2,5),\, \cU(2,5),\, \cU(-0.5,1) \right)$\revisionchange{ position,}{} in meters. 

In total, optimizations have been completed for 20 targets, each for the \gls{sis}, \gls{e2log}, and \gls{cop} objective\revisionchange{, in order to make the results statistically meaningful}{}. The individual cost functions evolutions, positional mean integral error norms, and estimation uncertainties are evaluated from this set of trajectories. \revisionchange{It needs to be mentioned that f}{F}or the positional mean integral error norm evaluation, we chose to randomly perturb the coefficients $k_f$ and $k_m$ in the ranges of $\pm\SI{1}{\percent}$ and $\pm\SI{5}{\percent}$ and fulfill 30 closed-loop flights, \revisionchange{always changing $k_f$ and $k_m$, for each trajectory}{changing them for every trajectory}. \revisionchange{Perturbations above $\pm\SI{5}{\percent}$}{Greater perturbations} are not considered as such a deviation from the nominal value might hint at problems at the parameter identification/estimation. The estimation uncertainty results are based on 10 runs of each trajectory with different randomly wrong initial guesses ($\pm\SI{30}{\percent}$) from ground truth ($k_f = \SI{3.375e-04}{\newton\per{\second^{-2}}}$ and $k_m = \SI{0.016}{\meter}$).

\subsection{Discussion}

\subsubsection{Cost Function Behavior}
\label{sec:cost_behavior}

We recorded each objective function's cost value ($F_{\text{INIT}}^{}$, $F_{\text{\gls{sis}}}^{}$, $F_{\text{\gls{e2log}}}^{}$, $F_{\text{\gls{cop}}}^{}$) at all iteration steps during each optimization run to evaluate the behavior of the costs by averaging all runs.

As \gls{e2log} and \gls{sis} have different orders of magnitudes, a normalization with the initial cost value was performed. \revisionchange{In addition, to ensure that the values of \gls{e2log} (potentially growing unbound) remain in a comparable range, we depict its inverse value.}{In addition, we chose to use the inverted values of \gls{e2log} as they can grow unbound and allows for better comparisons.} A decrease ($<$1) means an improvement, while an increase ($>$1) indicates a decline in the performance of the respective objective. As the optimization uses a gradient-free method, the average shows some spikes\revisionchange{; this can be addressed in future works by adopting a gradient-based approach}{}.

\figref{convergence} presents the results of the 20 individual \gls{cop} runs.
The top left plot shows the average and $1\sigma$ standard deviation of the \gls{sis}-based optimization, \secref{sis}, with its cost in blue and the \gls{e2log}'s cost in orange. One can see the decrease in the sensitivity cost, meaning that the state and input sensitivities are minimized (as expected). This comes, however, at the expense of the overall quality of observability, indicated by the increase of the inverse \gls{e2log} cost. Therefore, these results seem to indicate that \gls{sis} and \gls{e2log} can be two \emph{conflicting objectives}. 
On the top right are both costs, again blue \gls{sis} and orange \gls{e2log}, depicted for the quality of observability optimization, \secref{individual}. As we chose the inverse here, a decrease is equal to an improvement of the \gls{e2log}. Once again, we see the behavior of the left plot reflected in this optimization as well.
From those two plots one can infer that if one improves the other might get worse.
The two plots at the bottom of \figref{convergence} are the results of the optimization runs based on the \gls{cop} objective, \secref{cop}. We can see that even distributed weights, as described in \secref{cop}, can still result in slightly skewed solutions caused by the approximation of the individual utopia and nadir points $F_{i}^{O}$ and $F_{i}^{N}$. The solutions themselves are Pareto optimal, and the straightforward filtering ensures an overall decrease of all considered objectives. This indicates that \gls{cop} can balance and decrease all objectives.

\subsubsection{Control Tracking Error}

\begin{figure}[t]   
    \centering
    \vspace{1mm}
    \includegraphics[width=0.48\columnwidth]{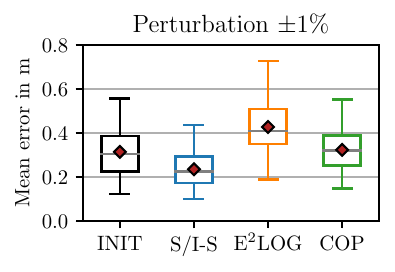}
    \hfill
    \includegraphics[width=0.48\columnwidth]{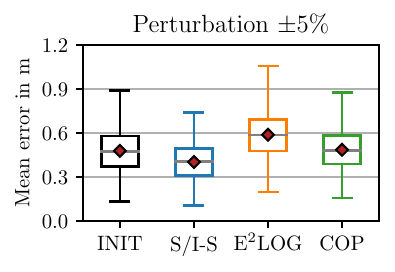}
    \caption{Quartile box plots showing the positional mean integral error norm over the whole trajectory, average over 20 trajectories each with 30 simulated closed-loop flights. The two plots show the influence of different perturbation amplitudes on the parameters $k_f$ and $k_m$. As expected, the \gls{sis} optimized one performs best followed by the \gls{cop}. The \gls{e2log} based trajectories perform worse as they just improve estimation performance. \gls{cop} as well as \gls{sis} are most effective with small perturbations, because the sensitivity is evaluated at $\vp = \vp_c$.}
    \label{fig:tracking} 
    \vspace{-1mm}
\end{figure}

The evaluation of the tracking performance of the system with its controller is based on the aforementioned set of trajectories, and is done by considering the positional mean integral error norm of position $\bm{r}(t)$ with respect to the desired position $\bm{r}_d(t)$.

The results can be seen in \figref{tracking} where we depict the quartile box plots from the statistical data. As mentioned before, each trajectory is flown in simulation 30 times with a changed set of $k_f$ and $k_m$ for each flight, normally distributed around $\pm\SI{1}{\percent}$ and $\pm\SI{5}{\percent}$ of their nominal values (which is used to evaluate the various sensitivity quantities).

The quartile boxplots of \figref{tracking} show the average tracking performance of each trajectory with different amplitudes of perturbation represented by the mean of the integral of the error norm at each point on the trajectory. The median of the \gls{sis} optimized trajectory performs the best and the \gls{e2log} one the worst, with \gls{cop} performing between those two, which confirms \figref{convergence} and our expectations. \gls{cop} can not reach the same performance level as \gls{sis} because of the balancing in \equref{cop_problem}, however, we can improve estimation performance at the same time, \figref{estimation}. 
Looking at the evolution of those boxplots, one can see that the farther away $\vp$ gets from $\vp_c$ the less difference is between each objective. This is also expected as the optimization evaluates the trajectories at nominal value. \revisionchange{According to these insights}{Therefor}, one might start with estimation-aware trajectories to get $\vp$ as close to $\vp_c$ and then switches to control-aware ones for improved tracking.

\subsubsection{Estimation Error}

\begin{figure}[b]   
    \centering
    \includegraphics[width=0.48\columnwidth]{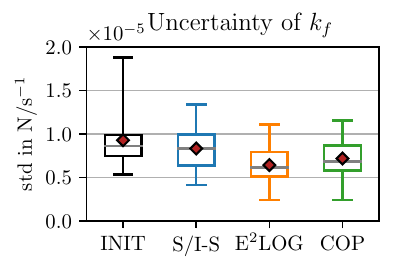}
    \hfill
    \includegraphics[width=0.48\columnwidth]{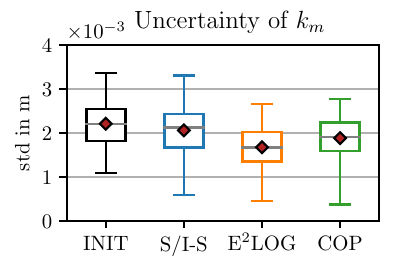}
    \caption{Quartile box plots showing the \gls{iekf}'s uncertainty based on the state's standard deviation at the end of the trajectory, average over 20 trajectories (initial, \gls{sis} optimized, \gls{e2log} optimized, and \gls{e2log} optimized each) closed-loop flights. Each trajectory is tested with 10 different initial guesses $\pm\SI{30}{\percent}$ uniformly distributed around ground truth. (left) is the thrust force coefficient $k_f$ and (right) the drag moment coefficient $k_m$ depicted. As expected, the initial trajectory has the worst estimation performance and the \gls{e2log} ones are the best with \gls{cop} generated motions being comparable.}
    \label{fig:estimation} 
\end{figure}

We used the \gls{iekf} implementation in Matlab of \cite{boehm_filter_2021} to evaluate the influence of the trajectories on the estimation of $k_f$ and $k_m$, which has proven to be good at estimating those parameters. The trajectory optimization is used to generate artificial position sensor and IMU measurements together with rotor speed input for each trajectory. Each of these recordings has been tested using the \gls{iekf} with initial guesses of nominal values $k_f$ and $k_m$ perturbed randomly by $\pm\SI{30}{\percent}$ 10 times. Note that both parameters are poorly observable. To visualize the influence, we once again use quartile boxplots for each individual optimization objective. \figref{estimation} shows on the left plot the quartile boxplot of $k_f$ and on the right for $k_m$. The estimation performance is evaluated by the reduction of uncertainty (represented by the standard deviation) at the end of the trajectory at $T$.

One can see that the \gls{sis} optimized trajectories perform slightly better than the initial ones. This is because we already gain more motion and excitation from the optimization objective. As expected, based on the optimization data, the \gls{e2log} optimized trajectories perform best with the new \gls{cop} optimized one in between. Note that this is expected, due to the balancing of two objectives we are not able to perform as well as a single objective optimized trajectory. These comparable results were already indicated in \figref{convergence}.
All the presented results support our claim that our proposed \gls{cop} objective can balance two \revisionchange{opposing}{\emph{possibly opposing}} objectives and maintain good performance. 

\subsubsection{Computation Times}
\label{sec:times}
All trajectory optimizations were done on a PC with an AMD Ryzen 5 3600 CPU (6 cores/12 threads), 16GB Ram, and NVidia RTX 2070 (Super) GPU. Note that one instance of the \gls{cop} implementation only uses one CPU thread, and therefore, parallelization is possible by starting multiple instances of \gls{cop}. The duration of each optimization run and each objective were logged. The initial trajectory generation is done in under $\SI{120}{\second}$. Minimizing the state and input sensitivity metrics ($F_{\Pi}^{}(\bm{a})$ and $F_{\Theta}^{}(\bm{a})$) takes on average $\SI{38}{\min}$ and  $\SI{15}{\min}$, respectively. Improving the observability through the \gls{e2log} ($F_{\text{E$^2$LOG}}^{}(\bm{a})$) needs on average $\SI{26}{\min}$. The trajectory optimization with \gls{sis} ($F_{\text{\gls{sis}}}^{}(\bm{a})$) finishes in $\SI{36}{\min}$ on average. The last objective of the \gls{cop} approach, minimizing both \gls{e2log} and \gls{sis}, runs for $\SI{27}{\min}$ on average. One complete optimization run needs in total around 2.4 hours which is due to the numerical integration of the complex metrics over the whole trajectory during each iteration of the optimization.


\section{Conclusion}
\label{sec:conclusion}
This paper wanted to answer the question: \emph{How to combine Closed-Loop State and Input Sensitivity-based with Observability-aware trajectory planning?} Our proposed Control \& Observability-aware Planning (COP) framework and its statistical evaluation provide an answer to it.

Intuitively, taking state and input sensitivities into account during the trajectory generation might result in non-informative trajectories for state/parameter estimation. However, informative motions for the estimation process are often difficult to control and likely cause higher tracking errors.
Our statistical case study of a 3D quadrotor \gls{uav}, focused on system parameters that have a significant impact on the tracking error and are poorly observable (thrust force coefficient $k_f$ and drag moment coefficient $k_m$), provided insights into the negative correlation between closed-loop state and input sensitivity-based and observability-aware trajectory planning.
We were able to show that both objectives work against each other, meaning that while one can improve, the other will get worse. 
Applying the Augmented Weighted Tchebycheff Method in a multi-step approach to such Multi-objective Optimization Problem (MOOP) allows to balance both in a Single-Objective Optimization Problem (SOOP), improving trajectory tracking and, at the same time, state estimation.

To summarize, we have successfully shown that it is important to \emph{consider both objectives} as they correlate with each other. The insights and results in this paper give a motivation to go forward with more sophisticated MOOP approaches. Further research should be conducted towards a real-world closed-loop flight using the estimation in the optimization as feedback.









\bibliographystyle{IEEEtran}
\typeout{}
\IEEEtriggeratref{19}
\bibliography{references}

\end{document}